\documentclass[journal]{IEEEtran}

\ifCLASSINFOpdf
\else
   \usepackage[dvips]{graphicx}
\fi
\usepackage{url}

\usepackage{graphicx}
\usepackage{amsmath}
\usepackage{booktabs}
\usepackage{amssymb}
\usepackage{color}
\usepackage{subfigure}
\begin{document}

\title{D3N: Bring the Power of Diffusion Model to Defect Detection}

\author{Xuyi~Yu
\thanks{Xuyi Yu is with the Institute of Artificial Intelligence and Robotics, Xi'an Jiaotong University, Xi'an, China.

Corresponding author: Xuyi Yu (yuxuyi@stu.xjtu.edu.cn).}   


}     

\markboth{IEEE Transactions}
{Shell \MakeLowercase{\textit{et al.}}: Bare Demo of IEEEtran.cls for IEEE Journals}
\maketitle

\begin{abstract}

Due to the high complexity and technical requirements of industrial production processes, surface defects will inevitably appear, which seriously affects the quality of products.
Although existing lightweight detection networks are highly efficient, they are susceptible to false or missed detection of non-salient defects due to the lack of semantic information. In contrast, the diffusion model can generate higher-order semantic representations in the denoising process. Therefore, the aim of this paper is to incorporate the higher-order modelling capability of the diffusion model into the detection model, so as to better assist in the classification and localization of difficult targets. First, the denoising diffusion probabilistic model (DDPM) is pre-trained to extract the features of denoising process to construct as a feature repository. In particular, to avoid the potential bottleneck of memory caused by the dataloader loading high-dimensional features, a residual convolutional variational auto-encoder (ResVAE) is designed to further compress the feature repository. The image is fed into both image backbone and feature repository for feature extraction and querying respectively. 
The queried latent features are reconstructed and filtered to obtain high-dimensional DDPM features.
A dynamic cross-fusion method is proposed to fully refine the contextual features of DDPM to optimize the detection model. Finally, we employ knowledge distillation to migrate the higher-order modelling capabilities back into the lightweight baseline model without additional efficiency cost. Experiment results demonstrate that our method achieves competitive results on several industrial datasets.

\end{abstract}


\IEEEpeerreviewmaketitle

\section{Introduction}

\IEEEPARstart{D}{efect} detection aims at finding out the location of defects from an image and assigning the labels of the categories. However, most of the state-of-the-art (SoTA) general object detectors \cite{sparsercnn,xu2022pp-yoloe,yolov5} are not designed for typical issues in defect detection. In addition, for complex defect detection in industrial scenarios, lightweight detectors are not capable due to limitations in their representational capabilities, while large-scale models cannot meet demands for real-time and strict resource constraints.

Many works \cite{kd1,kd2,kd3} have also proposed solutions such as distilling small models from larger ones, which require teachers with large scales (e.g., channels or depths). The semantic gap caused by the dimensionality gap between teacher and student makes it hard for the student to learn full knowledge of the teacher. In the process of knowledge distillation, there is a demand to design an adapter to adjust the channels if the number of channels differs between teacher and student, which leads to greater uncertainties in the learning between features at different semantic levels. Therefore, we want to achieve an extension of the network representation capability without changing the model dimensions, and apply the distillation technique to obtain more diverse and high-level features on a small model.

\begin{figure}[htbp]
\centering
\includegraphics[width=0.5\textwidth]{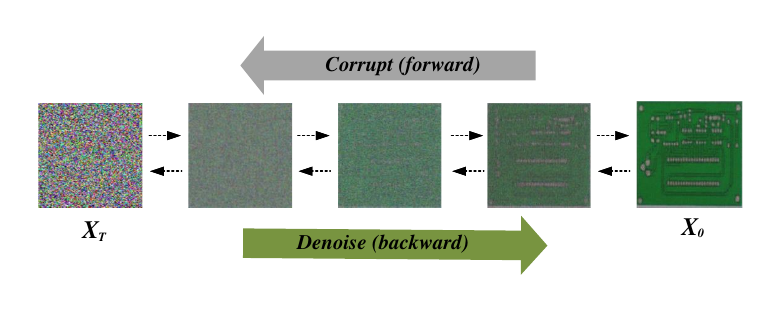}  
\caption{Diffusion models smoothly perturb data by adding noise, then reverse this process to generate new data from noise.}
\label{denoise}
\end{figure}

Denoising Diffusion Probabilistic Model (DDPM) \cite{ddpm} has recently received a lot of attention, crushing Generative Adversarial Network (GAN) \cite{gan} with its powerful generative capabilities. Due to its excellent performance, many works have also tried to introduce it into other tasks such as detection and segmentation \cite{graikos2022diffusion,wu2022medsegdiff,gu2022diffusioninst,ddpm-seg}. The intermediate activations in the backward denoising process of the diffusion model are shown to have higher-order semantic representations that can provide valuable information for downstream tasks. The perturbation and denoising process of the image in the diffusion model is shown in Fig. \ref{denoise}. Previous methods achieve the segmentation task based on the features from the diffusion denoising process, e.g., DDPM-Seg \cite{ddpm-seg}, MedSegDiff \cite{wu2022medsegdiff}, etc., and they construct a baseline to prove the effectiveness of the strategy, which provides us with valuable ideas. However, these methods need to perform the complete diffusion process, which leads to the unsatisfactory speed.

In order to combine the advantages of state-of-the-art detectors to increase the upper limit of accuracy, we employ an additional diffusion model as an aid to increase the feature diversity, thus compensating for the lack of higher-order semantic in the lightweight detectors.
Previous approaches need to load high-dimensional features with a dataloader, which is highly demanding on device memory. Methods of dimensionality reduction incur significant performance degradation, and features of the diffusion model may not be clearly differentiated on the channel. Therefore, we design a plug-and-play Residual convolutional Variational Auto-Encoder (ResVAE) for feature compression and subsequent recovery. Unlike previous methods of image-based reconstruction, we aim to reconstruct the higher dimensional DDPM features. In our method, we use images with labels and latent vectors as items in the dataloader and employ a pre-trained decoder to recover the high-dimensional DDPM features before loading them into the network. The proposed compression method provides a solution to the memory bottleneck in dataloader as an optional component.

As mentioned above, the denoising process of the diffusion model is relatively slow, which makes diffusion-based methods fail to meet real-time requirements. Therefore, this paper further proposes to leverage knowledge distillation to transfer the capabilities of hybrid models to existing state-of-the-art detection models to further improve the upper bound of performance , while not influencing the final inference architecture.

\begin{itemize}
\item[1)] Compared to state-of-the-art methods, we exploit the features extracted from the image backbone and diffusion model simultaneously. Specifically, features with higher-order semantics from the denoising process are employed to enhance the multi-scale prediction of the detection model. Experimental results on several industrial datasets demonstrate effectiveness.

\item[2)] Unlike previous methods that only load images with corresponding labels, this paper takes high-dimensional features from DDPM as loading items as well. A optional feature compression and recovery process is proposed, which makes it possible to reduce the memory requirements of the system dataloader under the premise of maintaining the integrity of the features.

\item[3)] We propose a dynamic cross-fusion method for dynamically assigning higher-order semantics of the diffusion model to the detection model. In order to avoid potential effects of noise, smoothing of high-frequency signals in the frequency domain is performed on the reconstructed DDPM features.

\item[4)] 
In summary, we employ a multi-stage strategy that injects higher-order features into the detector, with subsequent distillation based on the hybrid model. Our method avoids the cumbersome multi-stage process and retains the advantage of a lightweight baseline. Notice that we did not modify the dimensions in the model, and distillation between detectors of the same scale can achieve the optimal transfer of features.

\end{itemize}

\section{Related work}

\subsection{Diffusion Model}
Recently, diffusion models have become one of the hottest topics in the field of computer vision, demonstrating impressive generation capabilities. Compared to GAN \cite{gan}, diffusion models have an inherent drawback of requiring a large number of sampling steps and longer sampling time. Likewise, compared to variational autoencoders (VAEs) \cite{vae}, diffusion models also suffer from a large number of sampling steps and longer sampling time. The diffusion steps using the Markov chain kernel only require small perturbations, leading to extensive diffusion. Furthermore, operable models need an equal number of steps during the inference process, which means thousands of steps are necessary for sampling the random noise until it ultimately transforms into high-quality data similar to the prior data. The most significant drawback of diffusion models remains the need for performing multiple steps within the inference time to generate a single sample.

Despite the many drawbacks of the diffusion model, it is still widely used in a variety of generative tasks due to its impressive results and simple process of optimization. Additionally, the latent representations learned by diffusion models have been found to be useful in discrimination tasks, including image segmentation \cite{amit2021segdiff,diffseg2,wu2022medsegdiff,ddpm-seg}, classification \cite{ddpm-cls}, and anomaly detection \cite{ddpm-det}. Note that it is essential to skip the entire diffusion process in the inference stage of these tasks. This confirms the broad applicability of denoising diffusion models.

\begin{figure*}[t]
\includegraphics[width=\textwidth]{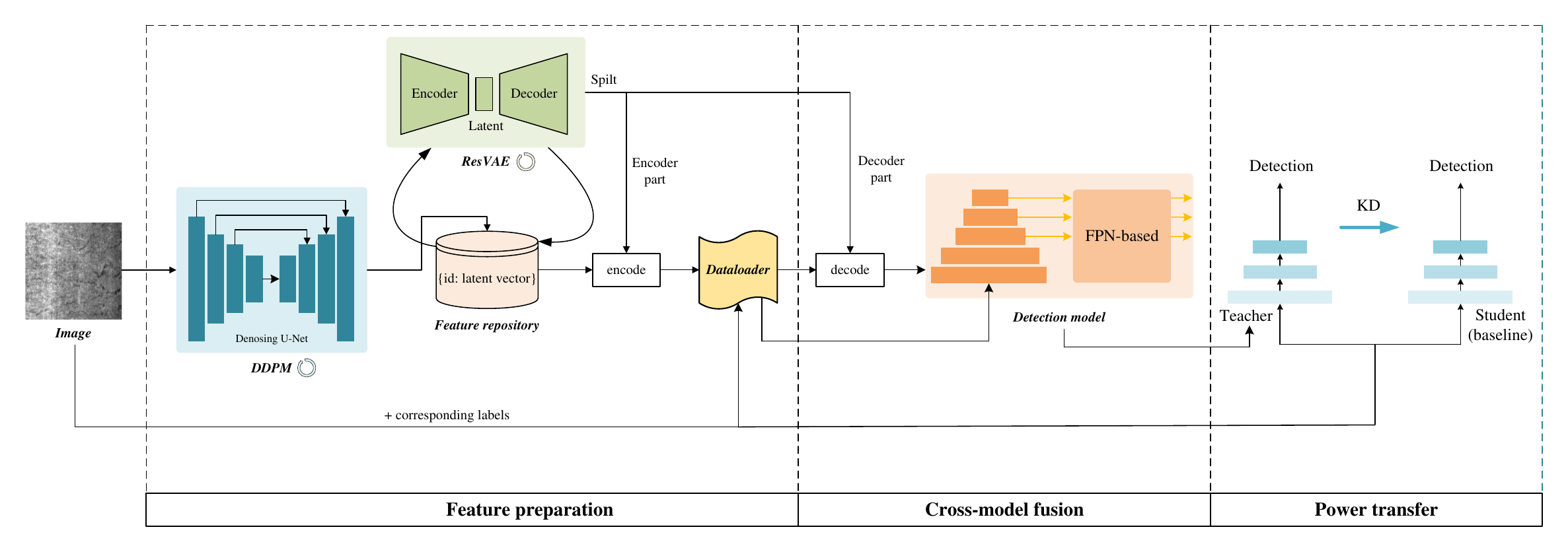}
\caption{The overall pipeline of the proposed method, which can be divided into three stages. The first stage is feature preparation, where a more condensed feature repository is constructed by unsupervised pre-training of the DDPM and VAE based on the input images and internal features respectively, where the circular arrows represent that the component needs to be pre-trained. The second stage is cross-model fusion, where the DDPM features from the repository are fused into the detection model for enhancement. The third stage is the power transfer, where the power of hybrid model is re-transferred to the baseline model.}
\label{fig1}
\end{figure*}

\subsection{Knowledge Distillation}
Large-scale deep models have achieved remarkable success, but their computational complexity and massive storage requirements make real-time deployment a challenge, especially on resource-constrained devices such as video surveillance and self-driving cars.

Knowledge distillation is a classical approach of model compression and acceleration that effectively learns small student models from large teacher models \cite{kd-1}. In knowledge distillation, small student models are usually supervised with the assistance of large teacher models, allowing the student models to mimic the teacher models for a competitive or even superior performance. Knowledge distillation is similar to the way humans learn, and the key issue in distillation is how to better transfer knowledge from large teacher models to small student models. Inspired by this, recent methods to knowledge distillation have been extended to teacher-student learning \cite{kd-2}, mutual learning \cite{kd-3}, self-learning \cite{kd-4} and so on. Most of the extensions to knowledge distillation focus on compressing deep neural networks. The generated lightweight student networks can be easily deployed in applications such as visual recognition, speech recognition and natural language processing (NLP).

In addition, the gap between the capacity of the teacher model and the student model affects the effectiveness of knowledge distillation, so the capacity of the teacher model needs to be controlled \cite{tea-stu}. We propose to build powerful teacher model without changing the capacity (width and depth) of the student model. This approach allows for optimal performance transfer between teacher and student.

\subsection{Defect Detection}
Similar to other vision tasks, the goal of defect detection is to locate and classify the targets from an image or video. Due to the difference between defects and natural objects, various methods \cite{es-net,df2-wacv,df1-cvpr} are proposed to solve the challenging problems in the industrial defect detection.

For the surface defect detection task of industrial products, ES-Net \cite{es-net} constructs a strong baseline and proposes several improvements to promote tiny defect detection, achieving state-of-the-art performance on multiple datasets.
A template-based detection method has been proposed for solving surface defects detection in fabrics and aluminum \cite{Reference-Based}, in which template-referencing and context-referencing algorithms are proposed for the problem of invisible background textures and the similarity between local detection boxes. However, template images need to be provided preliminarily, which is almost impractical in real-world scenarios.
A Foreground-perception Cycle-consistent Adversarial Network (FCGAN) \cite{df8} is proposed to recognize the foreground of samples and synthesize high-quality pseudo-defect images to improve the accuracy of defect detection from the perspective of data.
Tao $et~al.$ \cite{df3} implement Faster R-CNN in the cascading process to detect defects of insulators in power inspection. In the first stage, the insulators are located and cropped. Later, the defects are detected in the cropped region, and the whole network is trained end-to-end, but multi-stage approach leads to unsatisfactory efficiency.

Some methods have achieved state-of-the-art rankings in various datasets and competitions \cite{Yang2022MemSegAS,df6}, and they provide novel solutions for defect detection. However, the issue of efficiency is neglected due to more attention to accuracy.
In this paper, the features of the diffusion model are progressively refined and used to assist the detection model, and the capabilities of the obtained hybrid model are transferred to the baseline model by means of bootstrap learning, thus improving the performance of baseline while ensuring efficient inference.

\section{Proposed Method}
This section presents the details of the implementation of our method, which is mainly divided into three stages. The first stage is the feature preparation, in which the features with high-order semantics from the powerful DDPM are stored into the feature repository. The original feature repository is used as a data source, based on which the compressor is pre-trained.
The second stage is the cross-model feature fusion, in which the features in the feature repository are transferred to the feature fusion network of the detection model after remapping, noise filtering, and other operations. The third stage is power transfer, where we employ distillation techniques to re-transfer the capabilities of the hybrid model obtained in the second stage into a lightweight baseline model.

\subsection{Preliminary}
\subsubsection{YOLOv5 model}
YOLOv5 \cite{yolov5} is a most popular detection model, which has excellent trade-off between accuracy and efficiency. Compared to previous versions, YOLOv5 has been improved in terms of pre-processing, data augmentation, training optimization, and post-processing, etc. YOLOv5 provides different scales of the base model for different device requirements and application scenarios, which provides a high degree of flexibility. Similar to other general object detectors, it can be divided into three parts: the backbone network, the feature fusion network, and the final detection head. In the outputs of the detector, the bounding box is used for localization $(x, y, w, h)$, where $(x, y)$ are the center coordinates of the bounding box and $(w, h)$ are the width and height of the bounding box respectively. Additionally, preset category labels $(c)$ are used to classify each object.
\begin{figure*}[htb]
\centering
\includegraphics[width=\textwidth]{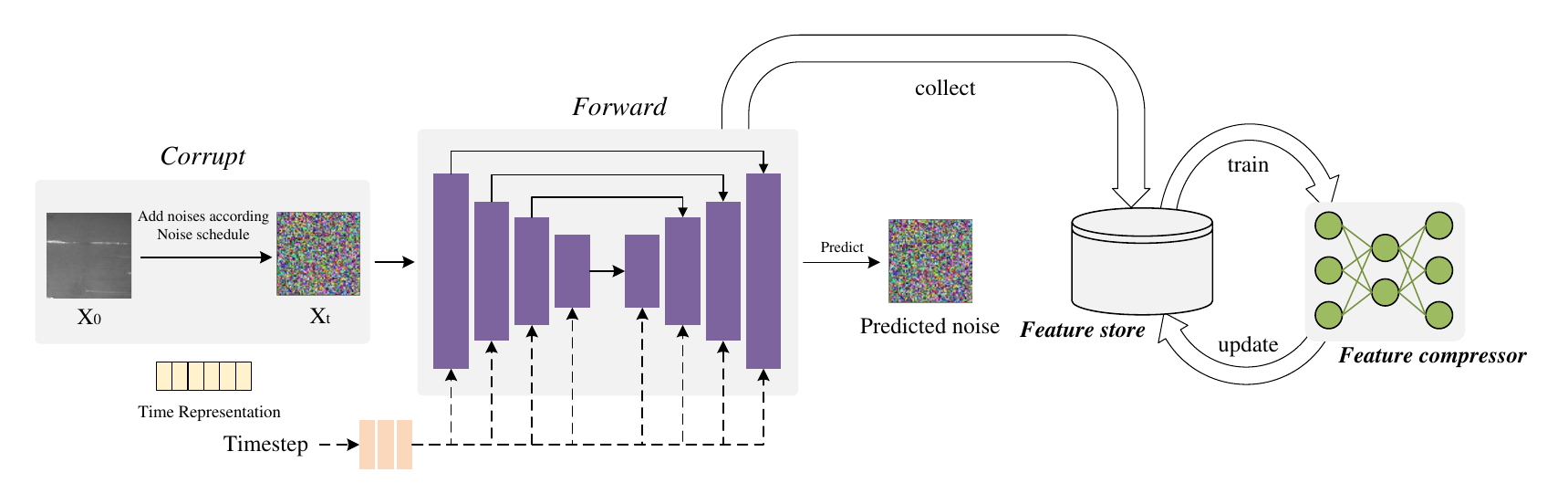}  
\caption{The process of feature preparation. Firstly, the input image $X_0$ is added with noise according to the noise schedule to get $X_t$. Then it is input to the pre-training DDPM for noise prediction, where the intermediate features of the UNet are collected in the feature repository. Finally, the feature repository is used as a data source to iteratively train and update the feature compressor (ResVAE) to obtain a more compact feature repository.}
\label{stage1}
\end{figure*}
\subsubsection{Diffusion Model}
The goal of the diffusion model is to convert a normal noise $X_T$ to $X_0$, where $X_t$ can be converted to $X_{t-1}$ with less noise by stepwise denoising.
Before that, the forward diffusion process requires gradual noise addition. $q(x_0)$ is the real distribution of data, and a real picture can be obtained by sampling from this distribution. The forward diffusion process is defined as $q(x_t|x_{t-1})$, which is the process of adding noise to the picture at each step, as shown in the following equation.

\begin{equation}
\begin{split}
q(x_t|x_{t-1})=N(x_t;\sqrt{1-\beta_t}x_{t-1},\beta_tI)
\end{split}
\end{equation}
where N denotes a normal distribution with mean and variance of $\sqrt{1-\beta_t}$ and $\beta_t$.
Alternatively, $x_t$ can be obtained directly from $x_0$ by theoretical conversion.
\begin{equation}
\begin{split}
q(x_t|x_0)=N(x_t;\sqrt{\hat{\alpha_t}}x_0,(1-\hat{\alpha_t})I), \\
x_t = \sqrt{\hat{\alpha_t}}x_0+\sqrt{1-\hat{\alpha_t}}\epsilon, \epsilon ~ N(0,1),
\end{split}
\end{equation}
where $\alpha_t = 1-\beta_t$,  $\hat{\alpha_t}=\prod^t_{s=1} \alpha_s$. 
The backward denoising process of the diffusion model can be fitted using a neural network.
\begin{equation}
\begin{split}
p_\theta(x_{t-1}|x_t)=N(x_{t-1};\mu_\theta(x_t,t),\sum_\theta(x_t,t)).
\end{split}
\end{equation}
The noise predictor network $\theta(x_t, t)$ predicts the noise component at the step t; the mean is then a linear combination of this noise component and $x_t$. The covariance predictor $\sum \theta(x_t, t)$ can be either a fixed set of scalar covariances or learned as well \cite{diff1}.
The denoising model $\theta(x_t, t)$ is typically parameterized by different variants of the UNet architecture.

\subsection{Feature preparation}
In previous work, features in DDPM are shown to contain higher-order semantics that can be employed for tasks such as semantic segmentation. Therefore, we try to leverage them to assist multi-scale detection due to the similarity of the tasks. 

First, we pretrain the DDPM based on the defect dataset to provide more suitable features in view of the domain specificity. Then, we input each image into DDPM to extract the features of decoder, and build them through several cycles to get a preliminary feature repository, where the sampling timestep $t=\{100, 150, 250\}$ is selected. It is worth noting that the upsampling of features to a unified scale in DDPM-Seg\cite{ddpm-seg} is designed to accommodate the subsequent segmentation task, while the feature pyramid network (FPN) structure in the detection task is inherently adapted to decoder-type architectures. Therefore, we retained the original scale for the subsequent matching of the detection model.

DDPM-Seg \cite{ddpm-seg} loads high-dimensional features directly into the network, which is likely to cause memory overflow problems on some devices with poor performance. Furthermore, dimensionality reduction on features will inevitably lead to information loss, which motivates us to consider a recoverable compression scheme. VAE is a well known tool for dimensionality reduction as well as an unsupervised learning method. We adopt ResVAE here to be more adapted to the field of image compression, where the residual block in ResNet is employed as the basic module to speed up the training.


First, the feature repository is employed as a data source to pre-train the VAE. Notice that we only construct the feature repository using features from the training set to avoid data leakage. Then, the pretrained VAE is employed to compress high-dimensional DDPM features to latent vectors with lower dimensions. After several cycles of compression, we obtain a smaller feature repository in which the compressed latent features are used as the basic units. The process of feature preparation is shown in Fig. \ref{stage1}.

When loading process of a batch is completed, the pre-training VAE will reconstruct these features and the results are used to guide the multi-scale fusion process of the detector. This manner allows the dataloader to avoid direct access to high-dimensional features, significantly reducing the memory requirements of device.

\begin{figure}[t]
\includegraphics[width=0.5\textwidth]{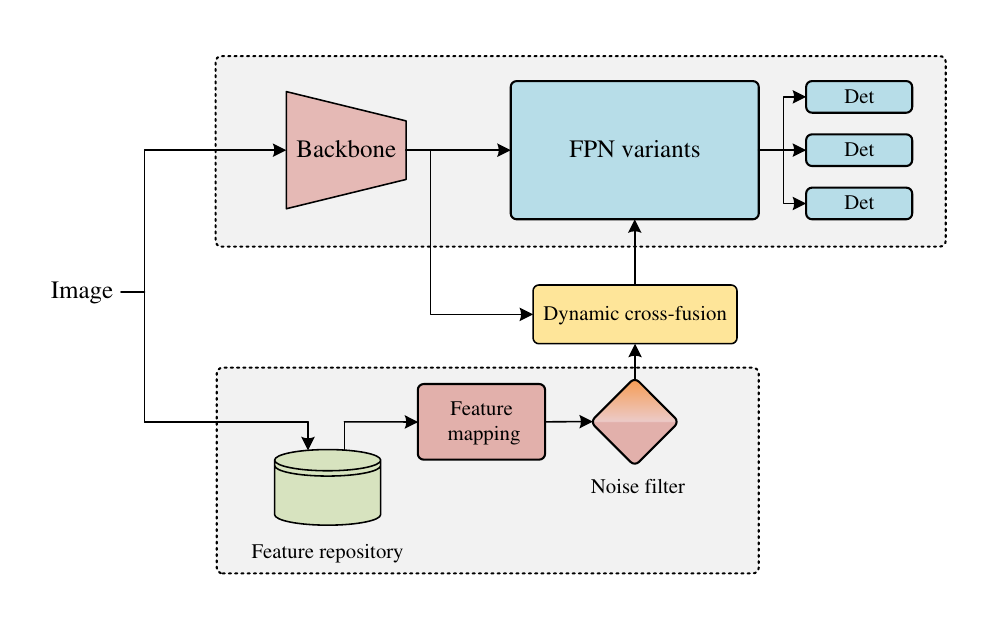}  
\caption{The structure of cross-model fusion. The input image is fed into both image backbone and feature repository to extract features respectively. The features from the feature repository are feature mapped to reconstruct the high-dimensional DDPM features. Subsequent noise filter is employed to further filter the high frequency noise from the features in the frequency domain. Finally, dynamic cross-fusion of features from different models is performed.}
\label{stage2}
\end{figure}

\subsection{Cross-model feature fusion}
\begin{figure}[t]
\centering
\includegraphics[width=0.5\textwidth]{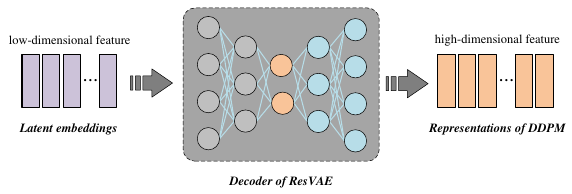}  
\caption{The process of feature mapping. A fixed-parameter decoder is used to convert latent features to high-dimensional features, where the grey part indicates no involvement in the forward process.}
\label{feature_map}
\end{figure}

The detection network can be generally divided into three parts, which are called the backbone network, the feature fusion network and the detection head, respectively. The backbone network is used to extract the multi-scale feature representation of the input image, and the feature fusion network is used to aggregate the multi-scale features, while the detection head uses the fused features to perform multi-scale prediction to obtain the final output of the model. Based on the overall flow of the detection model, we generated high semantics to assist the detection task using the previously constructed feature repository, and the proposed cross-model feature fusion is shown in Fig. \ref{stage2}.

For an image, we input it into the detection model, while taking it as a keyword to query in the feature repository to get the corresponding latent vectors. We then separate out the decoder part of the pre-trained VAE, which is employed to reconstruct the latent vectors to obtain the complete DDPM features. This process is called feature mapping, as shown in Fig. \ref{feature_map}, where the colored circles indicate that they are in effect during the inference.

In addition, since the input of the denoising process is a noisy image, the intermediate steps will inevitably be affected by noise. Therefore, inspired by MedSegDiff \cite{wu2022medsegdiff}, we transform the features to the frequency domain using the Fast Fourier Transform (FFT) and perform spatial attention in the frequency domain. Finally the features after attention are transformed back to the spatial domain by using inverse Fast Fourier Transform (iFFT). 



Subsequently, we employ cross-attention to fuse features from the DDPM into the detection model, achieving dynamic interactions between the different models. First, the input features $X \in \mathbb{R}^{H\times W \times C}$ and $Y \in \mathbb{R}^{H\times W \times C}$ are transformed by different linear transformations to obtain the Query ($Q$), Key ($K$) and Value ($V$), where $Q$ comes from $X$, $K$ and $V$ are from $Y$.
Self-attention is then calculated between $Q$ and $K$, and multi-head cross-attention is obtained by replicating in multiple subspaces. Cross-attention is implemented in the following equations, where a version with 4 heads is adopted.
\begin{equation}
\begin{split}
Q=Linear(X), 
K=Linear(Y),  
V=Linear(Y) \\ 
head = Attention(Q,K,V)  \\
MultiHead(Q,K,V)=Concat(head_1,...,head_4)   
\end{split}
\end{equation}

The single-head attention is calculated as shown in the following equation. 

\begin{equation}
\begin{split}
Attention(Q,K,V)=\frac{Softmax(QK^T)}{\sqrt{d_k}}V
\end{split}
\end{equation}

\begin{figure*}[t]
\centering
\includegraphics[width=0.8\textwidth]{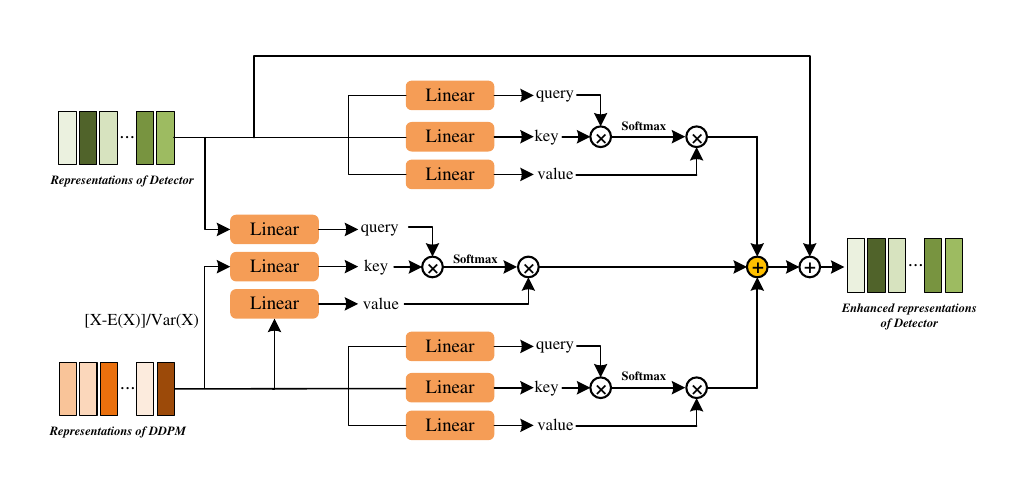}  
\caption{The structure of proposed dynamic cross-fusion operation. The reconstructed DDPM features and the features from detector are respectively performed with self-attention to smooth themselves, while both are also performed with cross-attention to achieve feature filtering of the DDPM features. Finally, three branches are fused in a learnable manner. Residual features from the detector are used for stable training.}
\label{cross-fusion}
\end{figure*}

It is worth noting that we take the features of the detector as query and the features of the DDPM for generating key and value, which can better extract the parts of the DDPM that are beneficial for the detection task.
In order to promote the network to perform a more flexible fusion of features from different models, we design a dynamic cross-fusion with multiple branches based on cross-attention, as shown in Fig. \ref{cross-fusion}. The DDPM features are normalized before performing cross-attention, and additionally self-attention is applied to each branch to further refine the features. Finally, these branches are fused in a learnable manner, which further helps the network to adaptively select effective features.

\subsection{Power Transferring}
After performing cross-model feature fusion, we make full use of DDPM features to enhance the detection model, but this comes at a large cost of efficiency.
With the aim to avoid the cumbersome process, we try to re-migrate the capabilities gained from the hybrid model to the baseline model. It is worth noting that our previous practices do not change the width and depth of the model, and the information transfer between models with same scale is the most efficient. Therefore, we employ knowledge distillation to perform power trasferring, i.e., using raw labels and hybrid models to co-optimize the baseline model.

We incorporate additional supervision from teachers for computing regression loss, objectiveness loss, and classification loss. In order to avoid learning the predictions of the teacher on the backgrounds, we represent the distillation loss as an object scaling function \cite{distill}. Only when the teacher predicts a high objectiveness value do we learn the border coordinates and category probabilities. Distillation loss is defined as follows.
\begin{equation}
\begin{split}
Loss = loss_{obj}^{distill}+loss_{cls}^{distill}+loss_{bbox}^{distill}
\end{split}
\end{equation}
\begin{equation}
\begin{split}
loss_{obj}^{distill} = loss_{obj}(o^{gt},o^s)+\lambda loss_{obj}(o^T,o^s) \\
loss_{cls}^{distill} = loss_{cls}(p^{gt},p^s)+ o^T \dot \lambda loss_{cls}(p^T,p^s) \\
loss_{bbox}^{distill} = loss_{bbox}(b^{gt},b^s)+o^T \dot \lambda loss_{bbox}(b^T,b^s)  \\
\end{split}
\end{equation}
where the $loss_{obj}^{distill}$, $loss_{cls}^{distill}$ and $loss_{bbox}^{distill}$ represent the distillation loss of objectiveness, classification and regression respectively. $o$, $p$, $b$ are the objectiveness, class probability and bounding box coordinates of the student network, and $o^{gt}$, $p^{gt}$, $b^{gt}$ are the values derived from the ground truth. The objectiveness is defined as Intersection of Union (IoU) between prediction and ground truth, class probabilities are the conditional probability of the category, the box coordinates are the normalized coordinates relative to the original image.

\begin{figure}[t]
\centering
\includegraphics[width=0.5\textwidth]{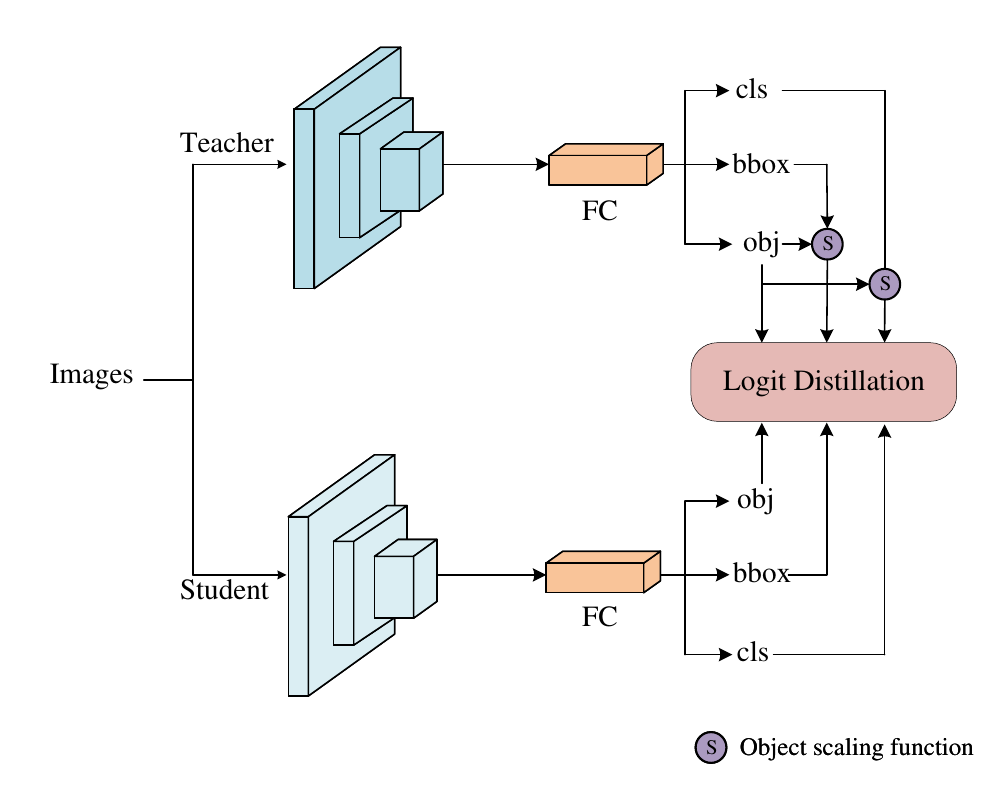}  
\caption{The process of power transferring, where the object scaling function denotes the use of objectiveness to scale the classification and regression branches. The predictions of the teacher are employed to guide the learning of student, while predictions from background are filtered out.}
\label{stage3}
\end{figure}

The augmented dataloader (including latent features, images, and corresponding labels) is used in training for knowledge transfer, while the vallina dataloader is used to test the student model to maintain consistency with the baseline.
The process of power transferring is shown in Fig. \ref{stage3}.

\begin{figure}[t]
\centering
\subfigure[crazing]{\includegraphics[width=0.3\columnwidth]{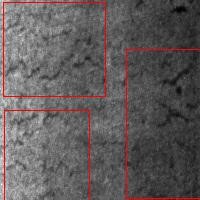}}
\subfigure[inclusion]{\includegraphics[width=0.3\columnwidth]{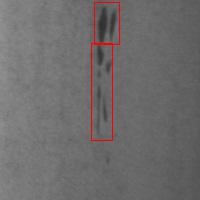}}
\subfigure[patches]{\includegraphics[width=0.3\columnwidth]{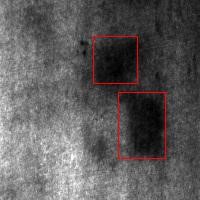}}

\subfigure[pitted\_surface]{\includegraphics[width= 0.3\columnwidth]{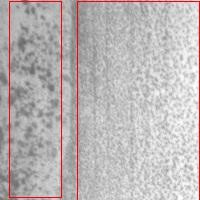}}
\subfigure[rolled-in\_scale]{\includegraphics[width=0.3\columnwidth]{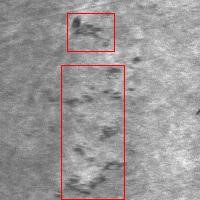}}
\subfigure[scratches]{\includegraphics[width=0.3\columnwidth]{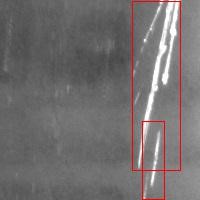}}
\caption{
Sample diagrams from each category of defect in the NEU-DET.} 
\label{fig7}
\end{figure}
\begin{figure*}[!t]
\centering
\includegraphics[width=0.8\textwidth]{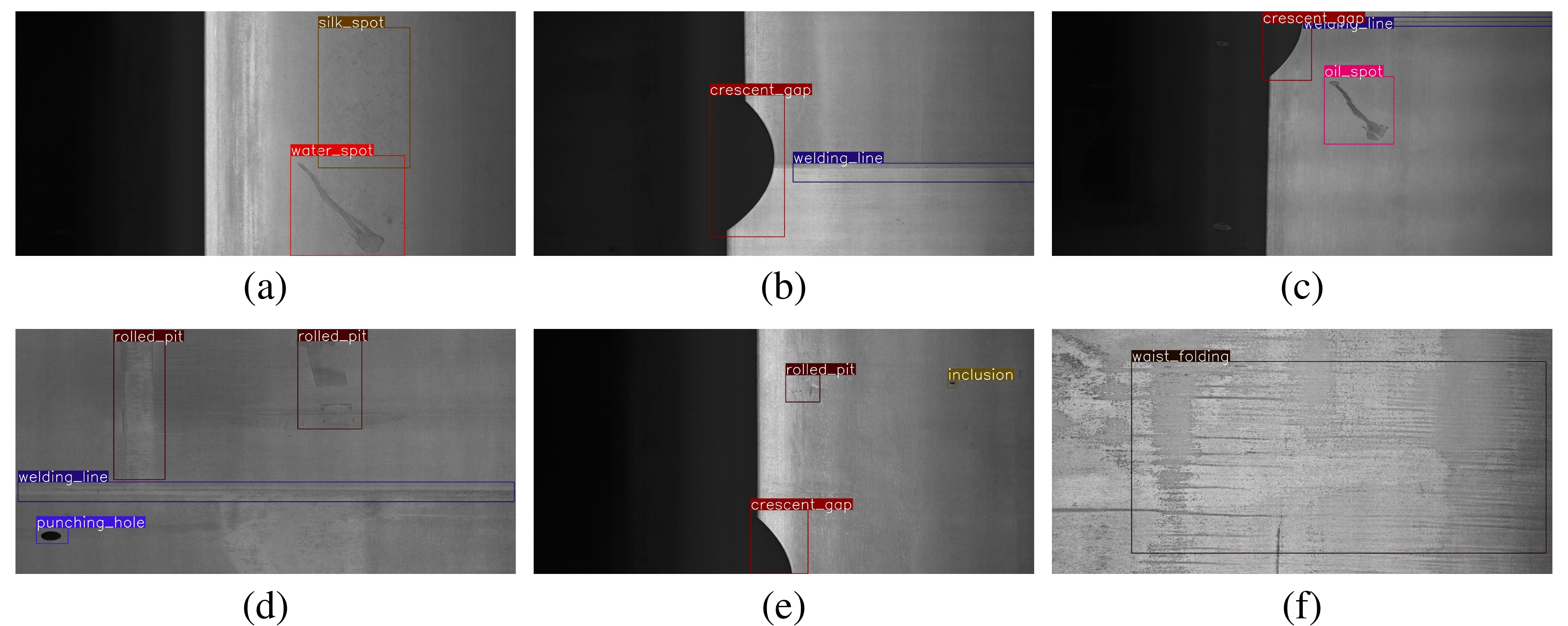}  
\caption{
Sample diagrams from each category of defect in the GC10-DET.} 
\label{fig8}
\end{figure*}

\begin{figure}[!t]
\centering
\includegraphics[width=0.5\textwidth]{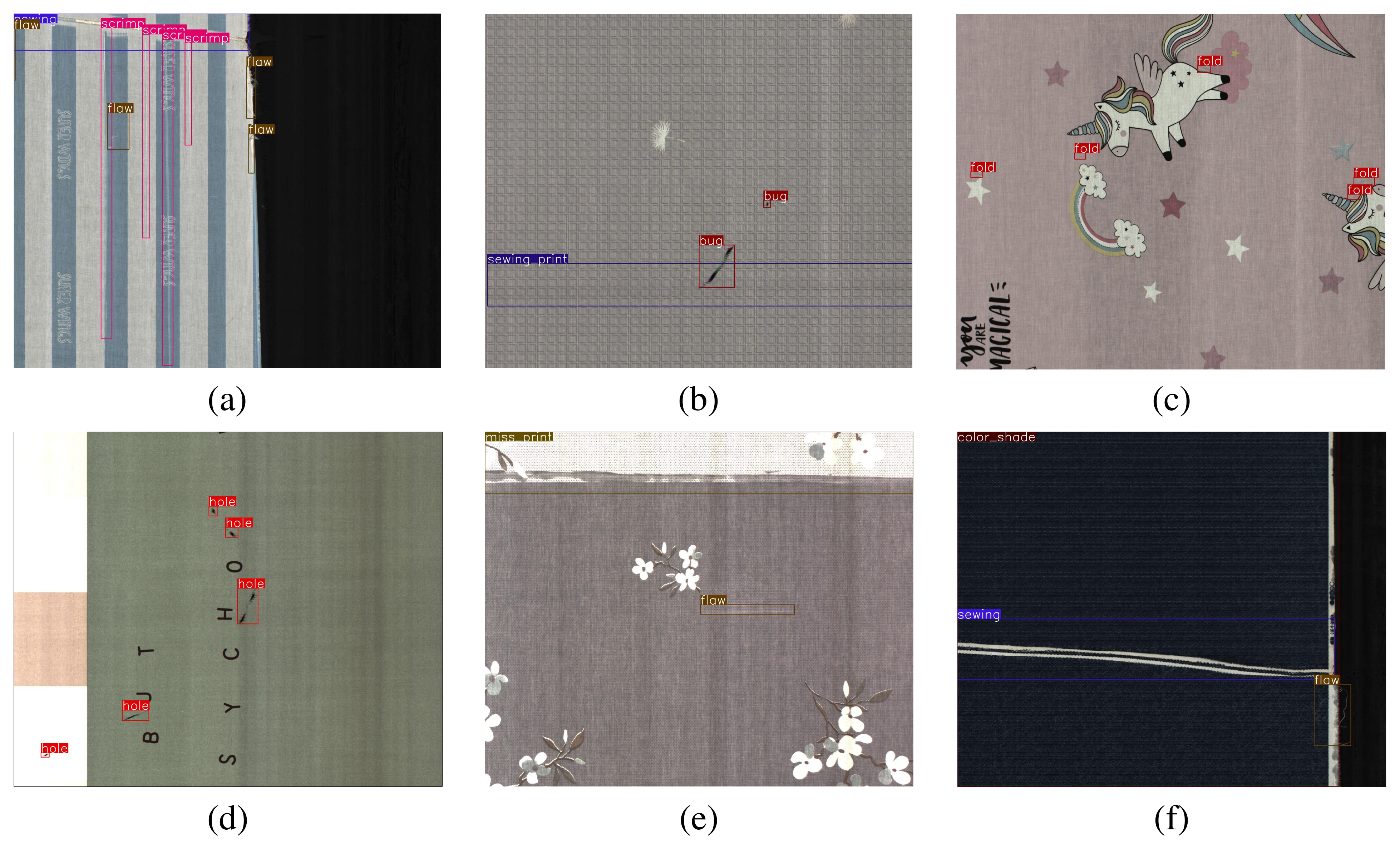}  
\caption{
Sample diagrams from each category of defect in the Tianchi fabric dataset.} 
\label{fig9}
\end{figure}

\section{Experimental Results and Analyses}

\subsection{Experimental Setup}
We use PyTorch to implement our method and experiment on two NVIDIA GeForce RTX 3090 GPUs. For DDPM, the total number of time steps is set to 1000, where the learning rate ranges from 5e-6 to 5e-8, and the corresponding learning rate schedule is cosine annealing. The Adam optimizer is used to optimize the parameters. In order to minimize the demand for graphic memory, we set the input image size to 64x64 and the batch size to 2. For ResVAE, we set the number of input channel to 512, the number of internal latent channel to 64, and the total training epoch to 2000. The KL (Kullback-Leibler) divergence loss and MSE (Mean Squared Error) is used to perform the loss computation, with adam as the optimizer. The KL penalty scale is set to 1 and batch size is set to 64. 
The configuration of the knowledge distillation is based on logit distillation, where the distance function adopts the L2 loss and the temperature coefficient is set to 20. We train the student model for 100 epochs with a initial learning rate of 0.01 and a batch size of 16. The student model is evaluated on the test set and report the accuracy. The configuration of the detector is consistent with YOLOv5.

For the case of inconsistent image aspect, instead of changing the padding strategy in DDPM to YOLOv5's letterboxing strategy, we scaled it directly. This is done for two reasons: one is to reduce the generation of black edges, and the other is to motivate YOLOv5 to learn the scaling information of a certain dimension.

We used the official library of YOLOv8 \cite{rath2023yolov8} to obtain the experimental results of YOLOv8 and RT-DETR. The results for the R-CNN family\cite{fasterrcnn,cascadercnn,sparsercnn}, ATSS \cite{atss}, and Autoassign \cite{zhu2020autoassign} models were obtained based on the mmdetection toolbox \cite{mmdetection}. However, training directly with the default configuration leads to a value of $Nan$ for the Hungarian matching loss, so we changed the optimizer of RT-DETR to stochastic gradient descent (SGD) and the corresponding initial learning rate to 5e-3.

\begin{figure*}[t]
\centering
\subfigure[Cross-Det-DDPM]{\includegraphics[width=0.3\textwidth]{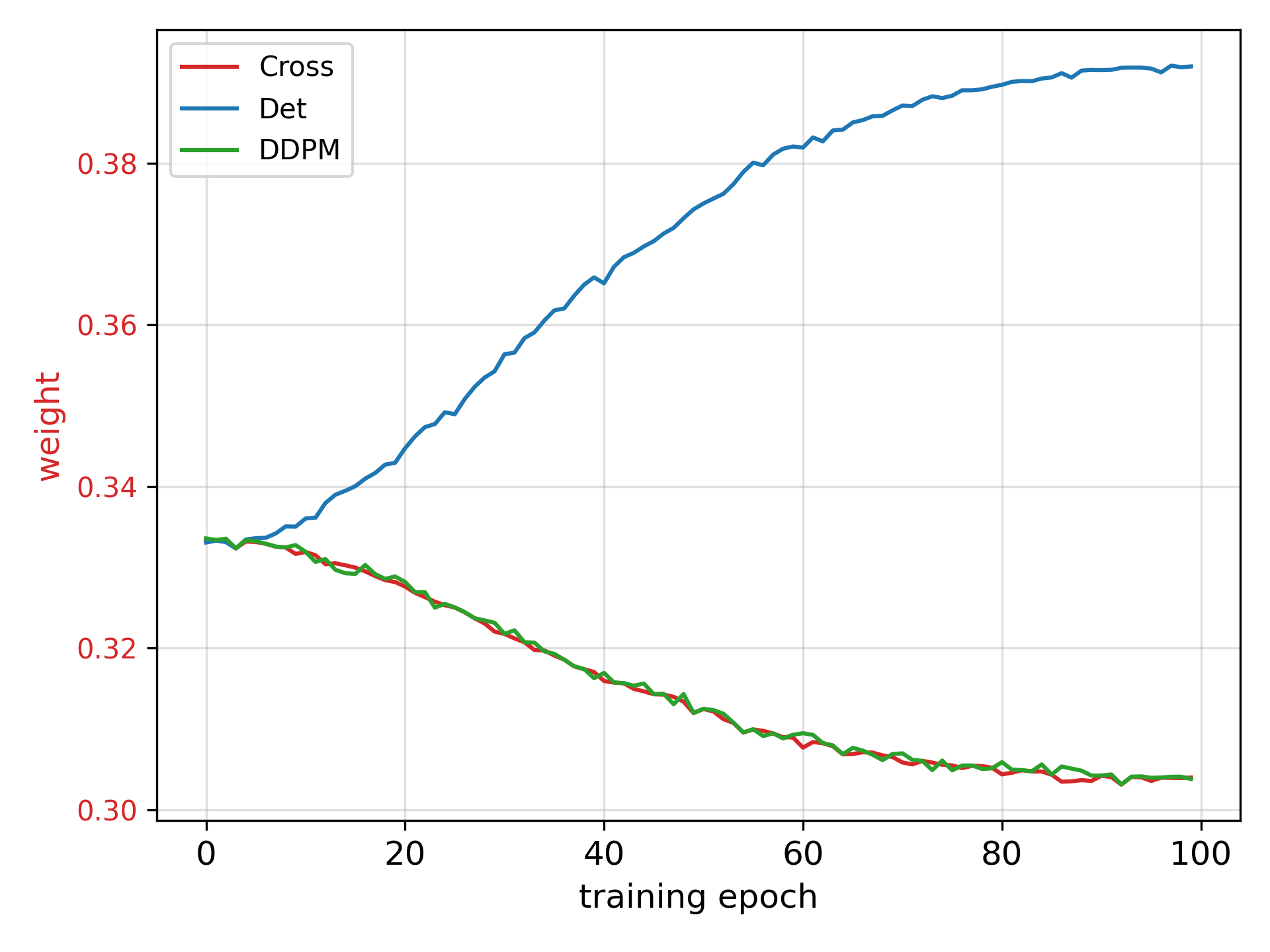}
\label{weightcurvea}
}
\subfigure[Cross-Det]{\includegraphics[width=0.3\textwidth]{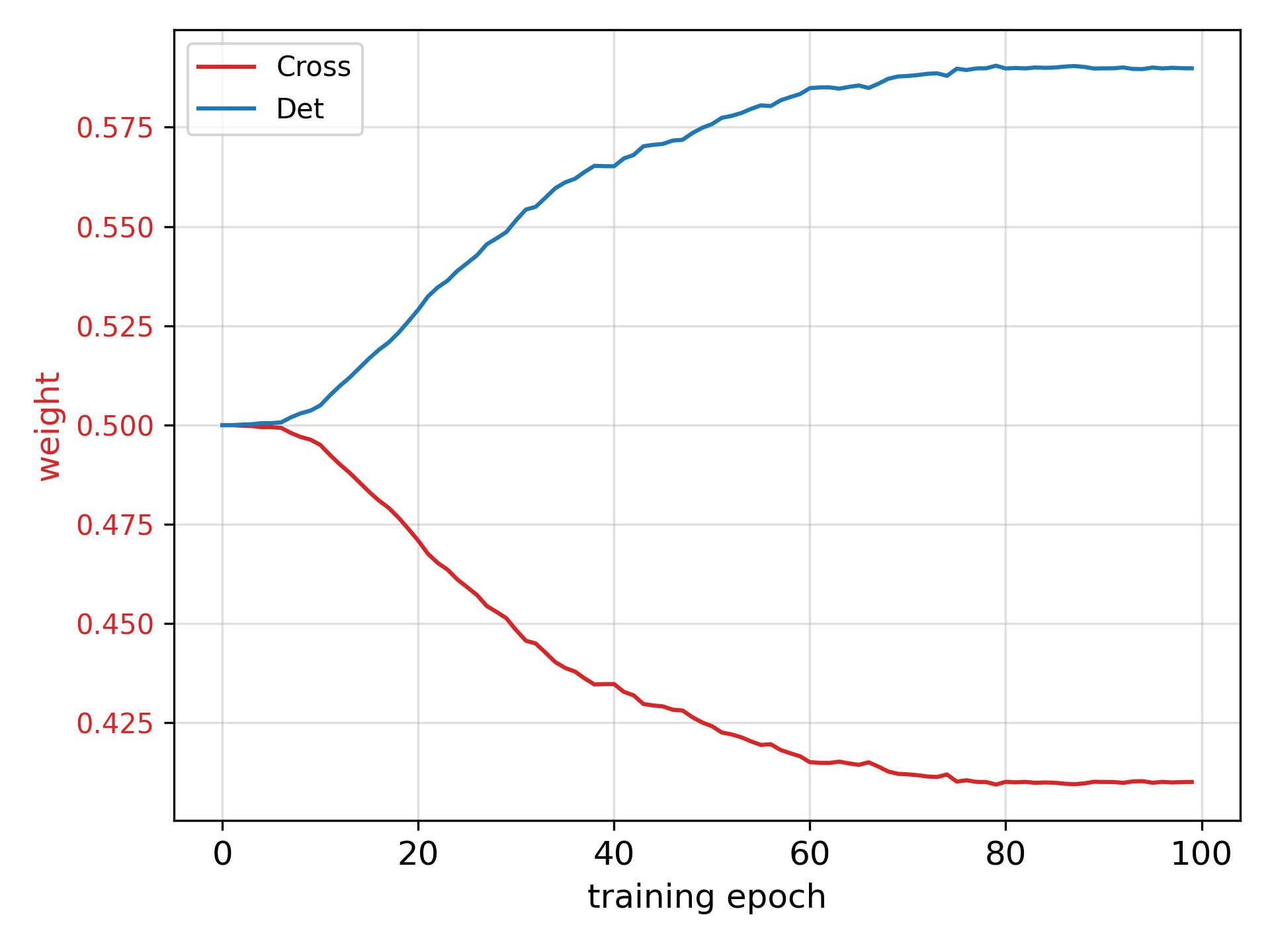}
\label{weightcurveb}}
\subfigure[DDPM-Det]{\includegraphics[width=0.3\textwidth]{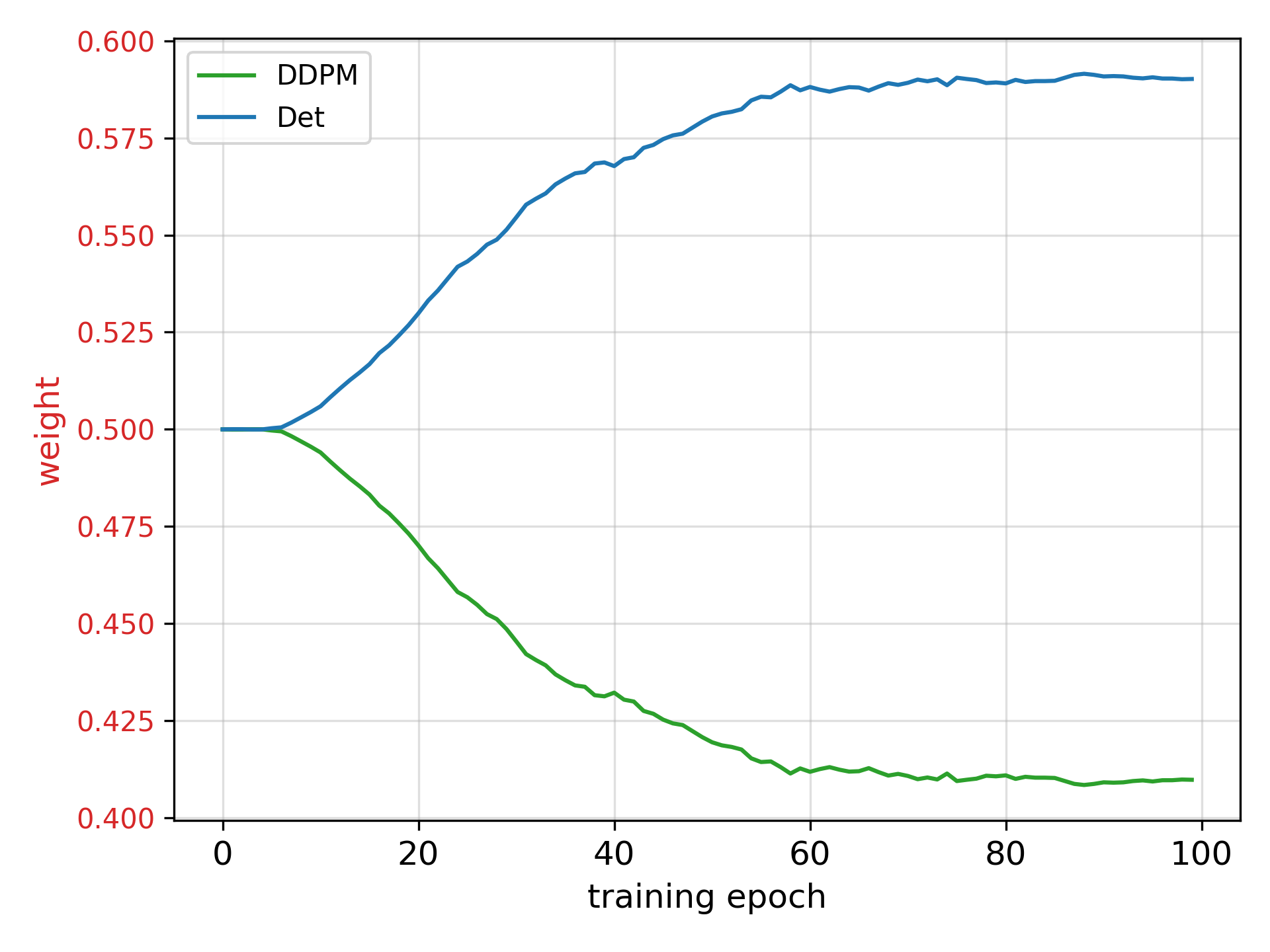}
\label{weightcurvec}}
\caption{The changes in the weight of each feature during training, where DDPM and Det represent the features of the DDPM and detection model, respectively, and Cross represents the cross-attention result between them. The figure (a) shows the changes in the weights of the three features, while (b) and (c) separately show the relative weights of the detection model feature and the other two features.}
\label{weightcurve}
\end{figure*}

\subsubsection{NEU-DET}

NEU-DET consists of 6 categories of defects in hot-rolled steel strips, including crazing, inclusion, patches, pitted surface, rolled-in scales, and scratches, and NEU-DET annotates each defect in VOC format. There are 300 images in each category, so there are 1800 images in total. Sample images of each category and corresponding annotation boxes in the NEU-DET dataset are shown in Fig. \ref{fig7}. Since the image resolution of the dataset is only 200×200 pixels, all images are scaled to 512×512 before being input to the network for training.

\subsubsection{GC10-DET}
GC10-DET is a dataset of metal surface defects collected in real industry. It contains ten types of surface defects, i.e., Punch (Pu), Weld (Wl), Crescent Gap (Cg), Water Spot Oil Spot (Os), Silk Spot (Ss), Inclusions (In), Rolled Pits (Rp), Creases (Cr), Waist Folds (Wf). The collected defects are on the surface of the steel plate. Some typical images containing all categories of defects are shown in Fig. \ref{fig8}. The dataset consists of 2294 grey images, and we constructed the training set and test set according to the ratio of 8:2.

\subsubsection{Tianchi Fabric Dataset}
The fabric defect dataset contains a total of 15 categories, and we choose 9 types of defects as targets in our experiments: sewing, sewing print, scrimp, bug, flaw, color shade, miss print, hole, and fold. The original images are sliced into two 2048×1696 pixels images, while the images without defects are removed. Finally, 4972 images are collected as a new dataset and divided according to 8:2. Typical images containing all the defects and the corresponding annotated boxes are shown in Fig. \ref{fig9}.


\subsection{Ablation Study}

\subsubsection{Contribution Analysis of DDPM and Detection Model}
To understand the contribution of each part intuitively, we visualize the fusion factors during training, which is shown in Fig. \ref{weightcurve}. As can be seen from the Fig. \ref{weightcurvea}, the proportion of detection features continues to be the highest, which is important for maintaining stability during training and ensuring final performance. When the optimization objective of the network is satisfied, there is no drastic change in the fusion factor, which indicates that each branch brings gains to some extent. To further quantify the importance of the two similar descent curves, we perform a weight visualization of the two branches separately, as shown in Fig. \ref{weightcurveb} and \ref{weightcurvec}. It can be seen that both DDPM and cross-attention features are incorporated into the detection network with smaller weights compared to the detection features. However, using only detection features also does not provide better accuracy performance than hybrid features, which also indicates that the fusion of various parts of the features is more beneficial for the detection task.


\begin{table}[t]
\centering
\caption{Ablation of components within a dynamic cross-fusion.}
\label{tabled_dynamic}
\begin{tabular}{ccccc}
\toprule[0.5pt]
  dual self-att  & weighted sum & residual & norm & mAP@0.5(\%) \\
    \midrule[0.5pt]
     -    & -  & -    & -   & 77.2  \\
     \checkmark      & -      & -  & -    & 76.3   \\
      \checkmark      & \checkmark      & -  & -    & 76.7  \\
    -    & \checkmark     & \checkmark  &   -   & 78.2    \\
\checkmark    & \checkmark     & \checkmark  &   -   & 78.6   \\
    
    \checkmark     & \checkmark     & \checkmark  & \checkmark     & 79.0   \\
\bottomrule[0.5pt]
\end{tabular}
\end{table}

\begin{figure*}[t]
\centering
\subfigure[mAP0.5]{\includegraphics[width=0.5\columnwidth]{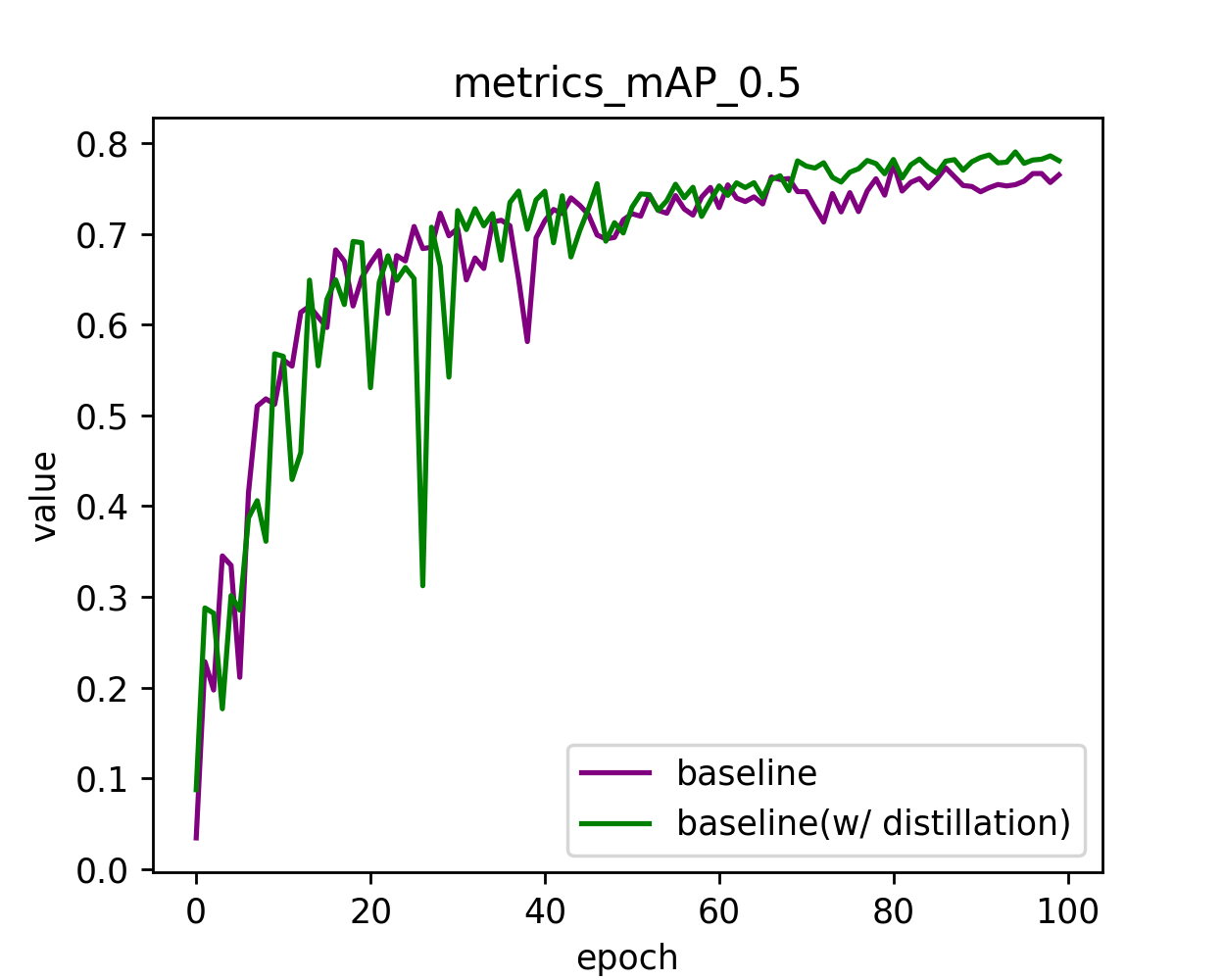}\label{mAP0.5}}
\subfigure[mAP0.5:0.95]{\includegraphics[width=0.5\columnwidth]{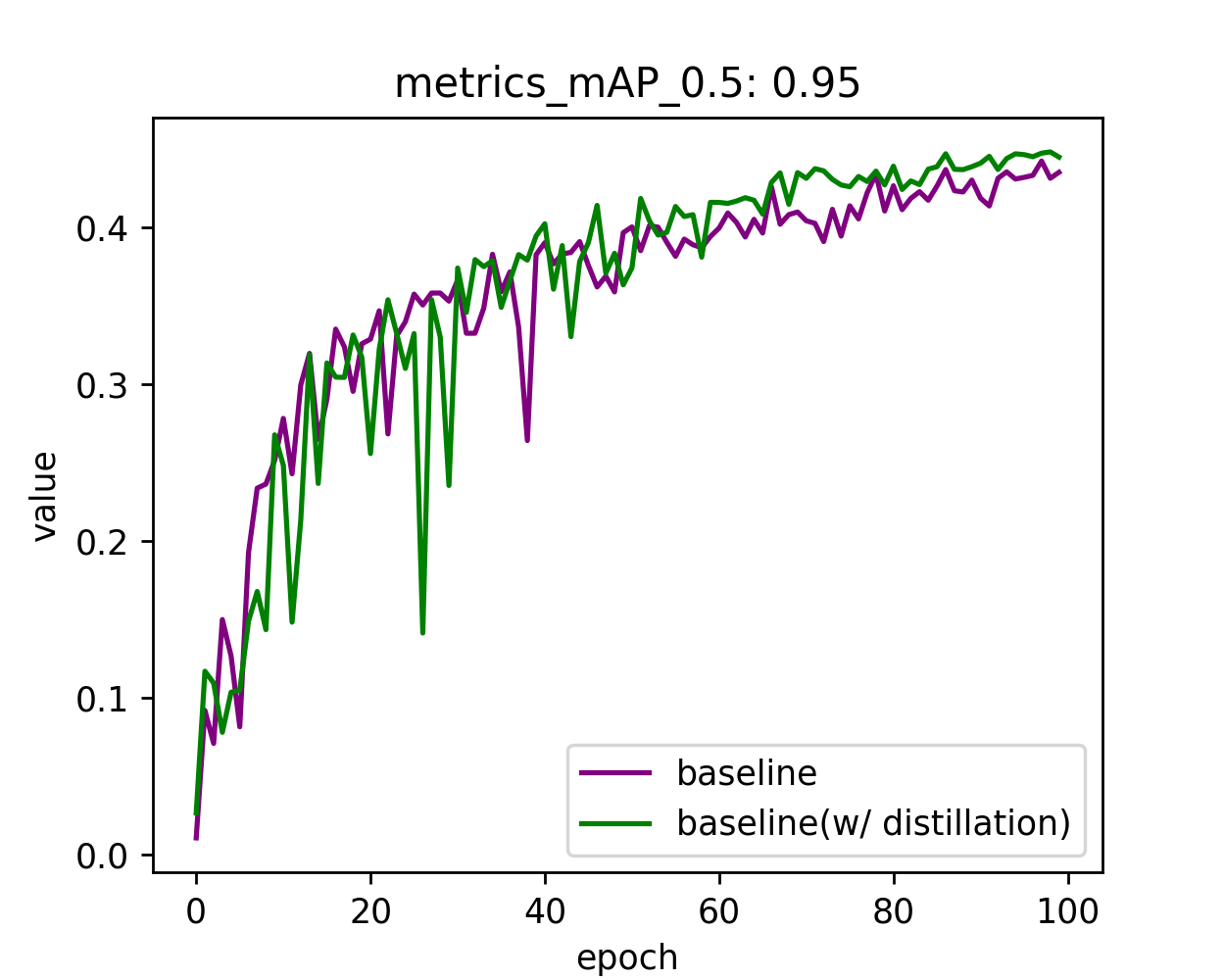}\label{mAP0.5:0.95}}
\subfigure[val\_box\_loss]{\includegraphics[width=0.5\columnwidth]{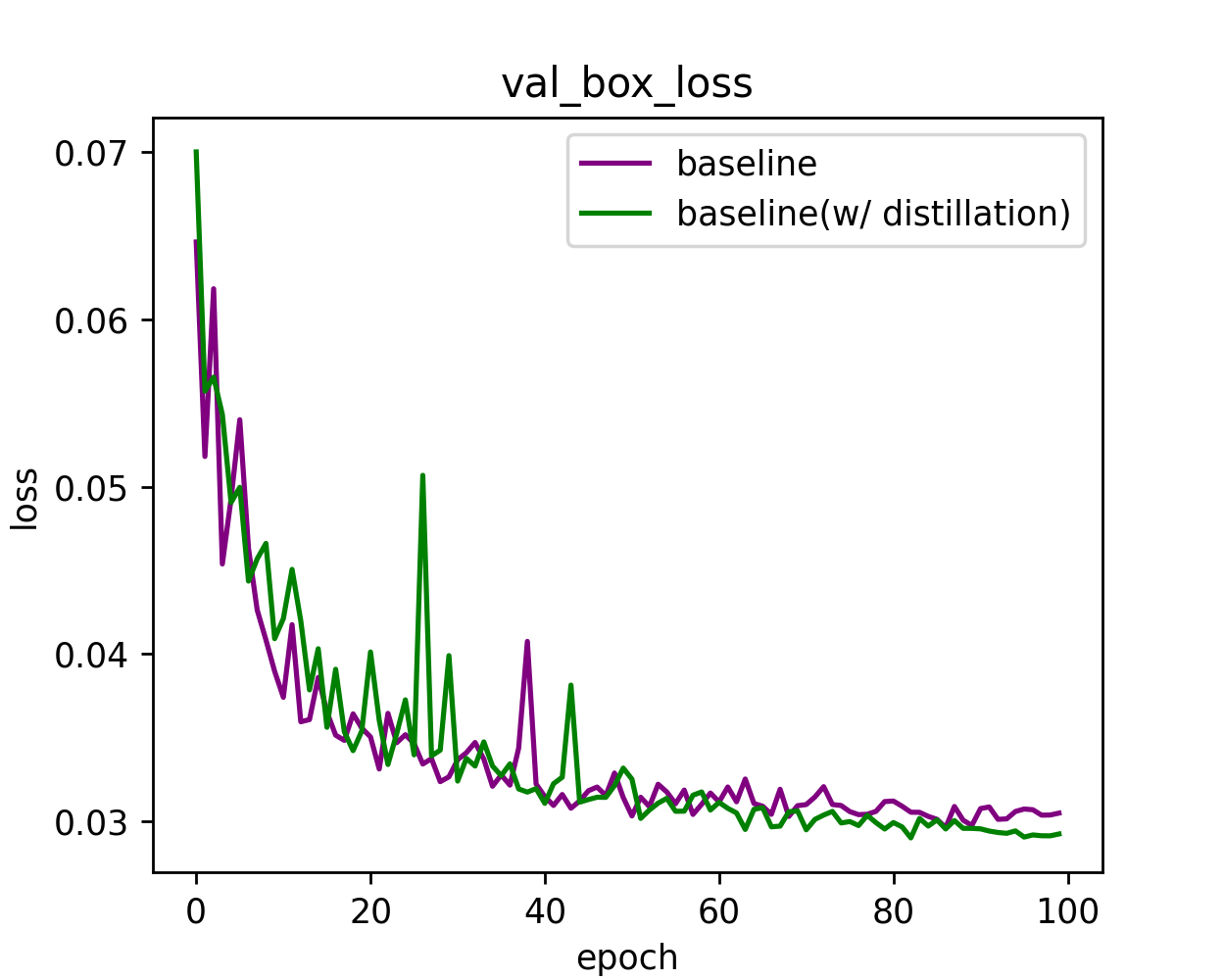}\label{val_box_loss}}
\subfigure[val\_obj\_loss]{\includegraphics[width=0.5\columnwidth]{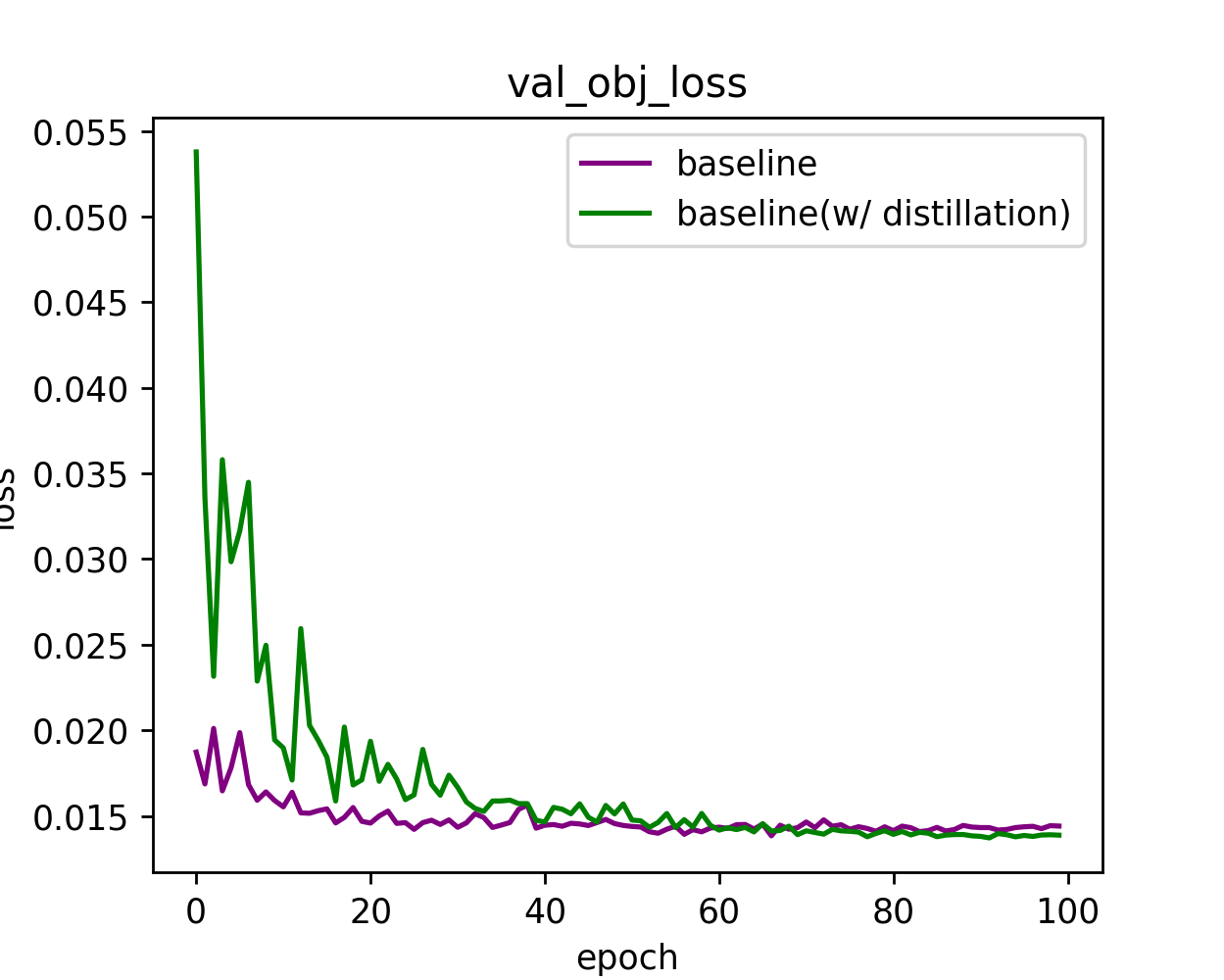}\label{val_obj_loss}}
\caption{Changes in the values of some metrics during training (mAP0.5, mAP0.5:0.95 and loss).} 
\label{metric}
\end{figure*}

\begin{table}[t]
\centering
\caption{Comparison of fusion methods.}
\tabcolsep=20pt
\label{table_fusion}
\begin{tabular}{cc}
\toprule[0.5pt]
Fusion operations  & mAP@0.5(\%)  \\
\midrule[0.5pt]
Concat + Conv1$\times $1  & 77.6          \\
Sum & 77.4            \\
Weighted sum   & 78.2          \\
Dynamic cross-fusion&  \textbf{79.0}         \\

\midrule[0.5pt]
\end{tabular}
\end{table}

\begin{table}[t]
\centering
\caption{Effect of each component on the baseline.}
\label{table_compare}
\begin{tabular}{cccccc}
\toprule[0.5pt]
PCA & ResVAE & Noise filter & Dynamic cross-fusion & mAP@0.5(\%) \\
\midrule[0.5pt]
\checkmark      & -         & -         & -  & 73.2   \\

-      &\checkmark     & -     & -   & 77.8    \\
-      &\checkmark      & \checkmark     & -   & 78.1    \\
\checkmark      &\checkmark     & \checkmark     & \checkmark   & 79.0 \\
\midrule[0.5pt]
\end{tabular}
\end{table}

\begin{table}[t]
\centering
\caption{Improvement of the proposed method on multi-task defect detection.}
\label{table_seg}
\setlength\tabcolsep{3pt}
\begin{tabular}{ccccccc}
\toprule[1pt]
Method   & P(Det) & R(Det) & mAP50(Det) & P(Seg) & R(Seg) & mAP50(Seg) \\
\midrule[0.5pt]
YOLOv5   & \textbf{0.829}  & 0.827  & 0.892  & \textbf{0.845}  & 0.793  & 0.858      \\
Proposed & 0.825  & \textbf{0.85}   &\textbf{ 0.901}  & 0.83  & \textbf{0.818} & \textbf{0.864}         \\    
\bottomrule[1pt]
\end{tabular}
\end{table}

\begin{table*}[t]
\setlength\tabcolsep{15pt}
\centering
\caption{Comparison of the test results (mAP, Param and FPS) of various methods and proposed method on the Tianchi fabric dataset.}
\label{tianchi}
\begin{tabular}{lcccccc}
\toprule[0.5pt]
Methods & Backbone &Image size & mAP@0.5 & Param  & FLOPs &FPS  \\
\midrule[0.5pt]
Faster R-CNN \cite{fasterrcnn} & ResNet-50 & 1333 $\times$ 800 & 60.8\%   & 315.4M  & 163G & 14  \\
Cascade R-CNN \cite{cascadercnn}& ResNet-50 & 1333 $\times$ 800 & 67.1\%   & 552.9M  & 191G & 11   \\
Sparse R-CNN \cite{sparsercnn}& ResNet-50& 1333 $\times$ 800 &    66.9\%    &    -    & - &    13  \\
RetinaNet \cite{focal}& ResNet-50 & 1333 $\times$ 800 & 64.5\%   & 246.5M  & 185G& 16   \\
ATSS \cite{atss} & ResNet-50 & 1333 $\times$ 800 & 66.1\%   & 246.6M  & 156G& 14   \\
AutoAssign \cite{zhu2020autoassign} & ResNet-50 & 1333 $\times$ 800 & 64.5\%   & 278.0M  & 156G& 12   \\
PAA \cite{paa}& ResNet-50 & 1333 $\times$ 800 & 65.3\%   & 246.7M  & 156G& 11   \\
PAA \cite{paa}& ResNet-101 & 1333 $\times$ 800 & 68.7\%   & 391.9M  & 215G& 9   \\
YOLOv3 \cite{redmon2018yolov3}& DarkNet-53 & 640 $\times$ 640 & 56.44\%  & 235.2M & 54.6G & 37   \\
YOLOv4 \cite{yolov4}& CSPDarkNet-53& 640 $\times$ 640 & 65.3\%   & 244.2M & 55.0G & 49   \\
YOLOv4-CSP \cite{scaled}& Modified CSPDarkNet & 640 $\times$ 640 & 68.0\%   & 200.4M  & 119.8G & 53 \\
ES-Net \cite{es-net} & Modified CSPDarkNet & 640 $\times$ 640& 76.2\%   & 147.98M& - & 56  \\
YOLOv5-s \cite{yolov5}& Modified CSPDarkNet v6 & 640 $\times$ 640 & 76.4\%  & \textbf{14.5M} & \textbf{16.5G} & \textbf{144}\\
YOLOv5-m \cite{yolov5}& Modified CSPDarkNet v6& 640 $\times$ 640 & 76.9\%   & 42.3M & 49.0G& 73 \\
YOLOv5-l \cite{yolov5}& Modified CSPDarkNet v6 & 640 $\times$ 640& 78.2\%  & 92.9M& 109.1G & 41 \\
YOLOv7 \cite{wang2023yolov7}& ELANNet & 640 $\times$ 640  & 75.8\%   &  74.8M& 104.7G & 45 \\
YOLOv8-s \cite{rath2023yolov8}& Modified CSPDarknet v7 & 640 $\times$ 640 & 73.9\%   &  22.5M & 28.6G& 123 \\
YOLOv8-m \cite{rath2023yolov8}& Modified CSPDarknet v7& 640 $\times$ 640  & 75.7\%   &  52.0M & 78.9G& 57 \\
YOLOv8-l \cite{rath2023yolov8}& Modified CSPDarknet v7 & 640 $\times$ 640 & 76.1\%   &  87.6M& 165.2G & 36 \\
RT-DETR-l \cite{RTdetr}& HGNetv2 & 640 $\times$ 640  & 76.7\%   &  66.2M & 110.0G& 84  \\   
Ours & Modified CSPDarkNet v6 & 640 $\times$ 640& \textbf{77.7\%} & \textbf{14.5M} & \textbf{16.5G}& \textbf{144} \\
\midrule[0.5pt]
\end{tabular}
\label{fabric_}
\end{table*}


\subsubsection{Ablation of the proposed dynamic cross-fusion}

We first perform an internal ablation of the proposed dynamic cross-fusion, which is shown in Table \ref{tabled_dynamic}. The group containing no components is simple cross-attention, which has a initial mAP of 77.2. Based on it, two-branch self-attention and weighted fusion are first performed, but they show a decrease in performance compared to the baseline, with a mAP of only 76.7. 
Then, a residual branch of the detection features is added, bringing the mAP to 78.2, an improvement of 1 percentage point from the baseline.
It is analyzed that self-attention is underutilized due to the lack of features from detectors to help stabilize training.
Then, two-branch self-attention with residuals is added and the mAP is further increased by 0.4. Finally, the DDPM features in cross-attention are normalized to further increase the upper limit of the mAP to 79.0.

We also compare the proposed dynamic cross-fusion with other common fusion operations, and the results are shown in the Table \ref{table_fusion}. It can be seen that the conventional Concat + 1$\times$1 convolution, summation, and weighted summation achieved results of 77.6, 77.4, and 78.2. We also tried to add various types of positional encoding for self-attention, but this resulted in worse performance. By introducing the designed dynamic multi-branch cross fusion, the mAP is further improved to 79.0. This indicates that our method learns a suitable fusion strategy that can extract the optimal features from each part to facilitate the detection task.

The knowledge of the final hybrid model will be distilled into the baseline, so the number of parameters and the amount of calculations (FLOPs) are actually consistent with the baseline and are not reported in the following tables.

\subsubsection{Ablation of the proposed components}
We performed ablation experiments on each component and the results are shown in Table \ref{table_compare}. First, a comparison of compression methods for the feature repository was conducted. It can be seen from the results that the dimensionality reduction of DDPM features using Principal Component Analysis (PCA) brings a negative effect on the baseline, with a mAP value of only 73.2, while VAE compression achieves better results. This indicates that DDPM features are not suitable for direct principal component extraction by channel, which will affect the original composition pattern and thus interfere with the learning of the detector. VAE is constructed in an encoder-decoder architecture that can recover the compressed features in a learnable manner, thus greatly preserving the integrity of the features.

Secondly, since the extracted features come from the denoising process of DDPM, there inevitably exists high-frequency noise. This problem can be well solved by the introduction of a noise filter, which is smoothed in the frequency domain, and the experimental results also demonstrate that the introduction of the noise filter brings an accuracy improvement of 0.4 mAP. Finally, the introduction of dynamic cross-fusion further increases the mAP of the detector to 79.0.

With all components integrated, we performed the distillation of the baseline model, as shown in Fig. \ref{metric}. It can be seen that the accuracy change of distillation has large fluctuation in the early stage, indicating that the guidance brought by the teacher and the ground truth is not consistent. In later stages, the network gradually learns the convergence of the two optimization directions to achieve a more stable accuracy improvement. As shown in Fig. \ref{val_box_loss} and \ref{val_obj_loss}, although the distillation model has a higher initial loss, we finally obtain a lower value of model loss by jointly optimizing the two objectives.

\begin{table}[t]\footnotesize
\setlength\tabcolsep{10pt}
\centering
\caption{Comparison of the test results (mAP) of various methods and proposed method on the NEU-DET.}
\label{neu-det}
\begin{tabular}{lcc}
\toprule[0.5pt]
Methods & Backbone & mAP@0.5  \\
\midrule[0.5pt]                
SSD \cite{liu2016ssd} & VGG-16   & 71.6\%  \\
Faster R-CNN \cite{fasterrcnn}& ResNet-34   & 70.0\%  \\
Faster R-CNN \cite{fasterrcnn}& ResNet-50   & 76.6\%  \\
DEA-RetinaNet \cite{cheng2020dearetinanet} & ResNet-50 & 79.1\% \\
CABF-FCOS \cite{yu2021cabffcos} & ResNet-50 & 76.7\% \\
DDN \cite{he2019ddn} & VGG-16   & 76.3\%  \\
DDN \cite{he2019ddn} & ResNet-50   & \textbf{82.1\%}  \\
YOLOv5s \cite{yolov5} & Modified CSPDarknet v6 & 76.6\% \\
YOLOv5m \cite{yolov5} & Modified CSPDarknet v6 & 77.2\% \\
YOLOv5l \cite{yolov5} & Modified CSPDarknet v6 & 77.3\% \\

ES-Net \cite{es-net} & CSPDarknet-53   & 79.1\% \\
YOLOv7-tiny \cite{wang2023yolov7} & ELANNet  & 72.0\%  \\
YOLOv7 \cite{wang2023yolov7} & ELANNet & 73.8\%  \\
YOLOX-s \cite{ge2021yolox} & Modified CSP v5 & 74.7\% \\  
YOLOX-m \cite{ge2021yolox} & Modified CSP v5 & 75.6\% \\    
YOLOX-l \cite{ge2021yolox} & Modified CSP v5 & 75.9\% \\      

YOLOv8-n \cite{rath2023yolov8} & Modified CSPDarknet v7 & 78.7\% \\ 
YOLOv8-s \cite{rath2023yolov8} & Modified CSPDarknet v7 & 77.8\% \\ 
YOLOv8-m \cite{rath2023yolov8} & Modified CSPDarknet v7 & 77.5\% \\    
YOLOv8-l \cite{rath2023yolov8} & Modified CSPDarknet v7  & 76.9\% \\    
RT-DETR-l \cite{RTdetr}  & HGNetv2 & 74.6\%  \\
Ours   &  Modified CSPDarknet v6  & 79.0\%  \\

\midrule[0.5pt]
\end{tabular}
\label{neu_}
\end{table}

\begin{table}[t]\footnotesize
\setlength\tabcolsep{10pt}
\centering
\caption{Comparison of the test results (mAP) of various methods and proposed method on the GC10-DET.}
\label{gc10}
\begin{tabular}{lcc}
\toprule[0.5pt]
Methods & Backbone & mAP@0.5  \\
\midrule[0.5pt]                

Faster R-CNN \cite{fasterrcnn} & ResNet-50   & 61.6\%  \\
Faster R-CNN \cite{fasterrcnn} & ResNet-101   & 62.8\%  \\
Cascade R-CNN \cite{cascadercnn} & ResNet-50   & 61.7\%  \\
Cascade R-CNN \cite{cascadercnn} & ResNet-101   & 63.9\%  \\
Dynamic R-CNN \cite{dynamicrcnn} & ResNet-101   & 64.5\%  \\
RetinaNet \cite{dynamicrcnn} & ResNet-50   & 66.6\%  \\
RetinaNet \cite{dynamicrcnn} & ResNet-101   & 66.3\%  \\

ATSS \cite{atss} & ResNet-50   & 64.9\%  \\
AutoAssign \cite{zhu2020autoassign} & ResNet-50   & 67.5\%  \\
PAA \cite{paa} & ResNet-50   & 66.6\%  \\
PAA \cite{paa} & ResNet-101   & 68.1\%  \\

YOLOv8-n \cite{rath2023yolov8} & Modified CSPDarknet v7   & 66.1\%  \\
YOLOv8-s \cite{rath2023yolov8} & Modified CSPDarknet v7   & 66.7\%  \\
YOLOv8-m \cite{rath2023yolov8} & Modified CSPDarknet v7   & 66.8\%  \\

Ours   &  Modified CSPDarknet v6  & \textbf{68.0\%}  \\

\midrule[0.5pt]
\end{tabular}
\end{table}

\begin{figure*}[t]
\centering
\subfigure[Tianchi fabric dataset]{\includegraphics[width=\columnwidth]{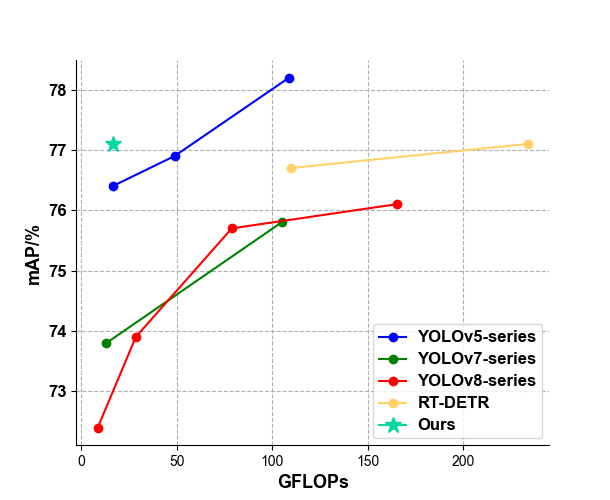}\label{compare_fabric}}
\subfigure[NEU-DET]{\includegraphics[width=0.9\columnwidth]{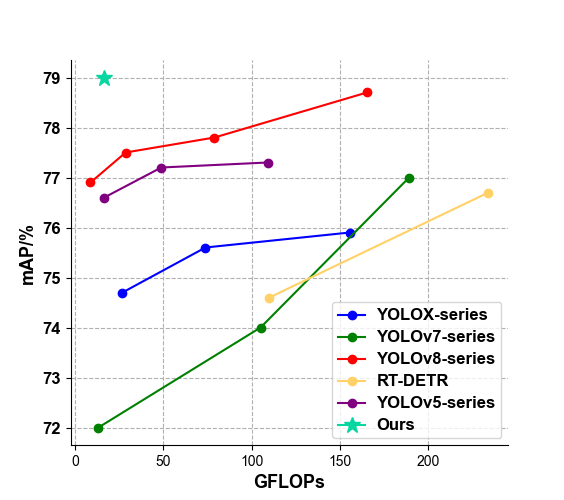}\label{compare_neu}}
\caption{Accuracy vs FLOPs. Left: Tianchi Fabrics dataset. Right: NEU-DET dataset. YOLOv5, YOLOX, RT-DETR and other series are added for comparison.} 
\label{line}
\end{figure*}




\begin{figure*}[t]
\centering
\subfigure{\includegraphics[width=2\columnwidth]{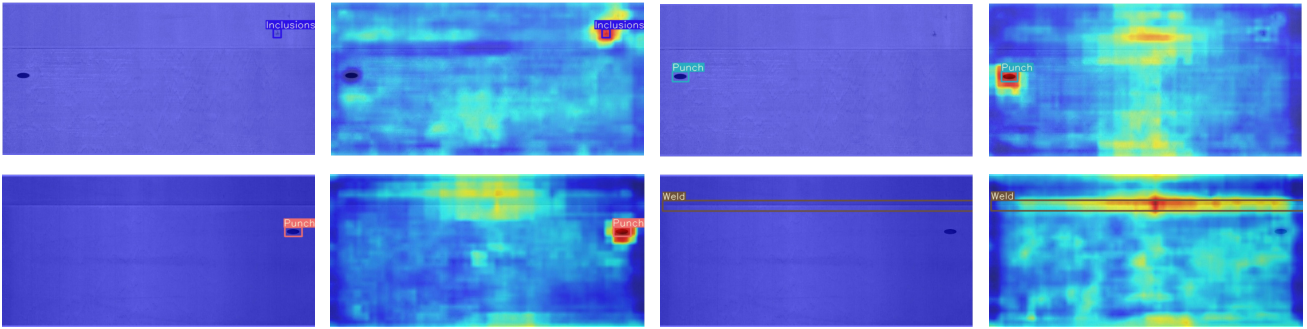}}
\caption{Visualization of the detection results on steel surfaces using GradCAM++, where each row represents the class activation results for different classes of defects in one image.} 
\label{heatmap}
\end{figure*}

\begin{figure*}[t]
\centering
\subfigure{\includegraphics[width=2\columnwidth]{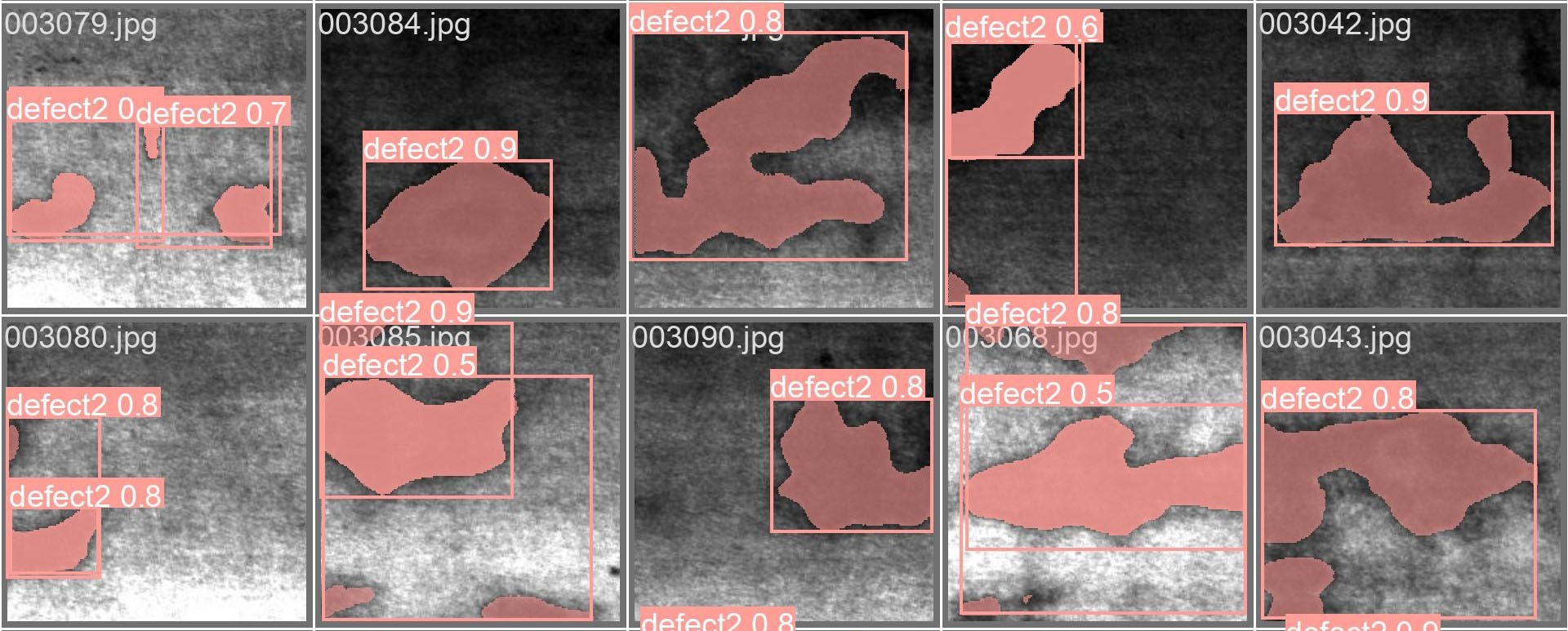}}

\caption{Some detection and segmentation results on the NEU-SEG dataset using the proposed method.} 

\label{vis4}
\end{figure*}

\subsubsection{Effectiveness of the our method in segmentation}
In order to validate the effectiveness of the proposed method on other tasks, the performance on multi-tasks (detection and segmentation) is tested as shown in Table \ref{table_seg}. The results show that our method achieves better performance in important segmentation metrics like recall and mAP, which suggests that stronger semantics are indeed effective in segmentation tasks.

\subsection{Comparison With the State-Of-The-Art Methods in Various Datasets}
In this section, our method is compared to benchmarks on several publicly available datasets and supplemented by testing some state-of-the-art object detectors such as YOLOv7 \cite{yolov7}, YOLOv8 \cite{rath2023yolov8}, RT-DETR \cite{RTdetr} and YOLOv5 \cite{yolov5} under local environment.

On the Tianchi fabric dataset, our method is superior in accuracy, while surpassing most of the detection methods in efficiency. Compared to two-stage R-CNN families such as Faster R-CNN \cite{fasterrcnn}, Cascade R-CNN \cite{cascadercnn} and Sparse R-CNN \cite{sparsercnn}, our method exceeds 16.9, 10.6 and 11.2 percentage points in mAP, respectively. For the single-stage methods RetinaNet and YOLO series (YOLOv3, v4, v5), our accuracy is improved by 21.3, 12.4, 1.3 percentage points, respectively. Meanwhile, for ES-Net, an improved method for the YOLO series, we have twice the speed and 1/10 the number of parameters of ES-Net \cite{es-net}, while the accuracy is comparable. For the latest SoTA detector RT-DETR \cite{RTdetr}, our method exceeds 1 mAP in accuracy while substantially superior in efficiency.

Then, the NEU-DET and GC10-DET are also used for evaluation. The test results of the proposed method and the benchmarks on the corresponding datasets are shown in Tables \ref{fabric_}-\ref{gc10}, respectively.

Fig. \ref{line} shows the performance comparison of the various methods on the Tianchi fabric dataset and NEU-DET. It can be seen that our method achieves the best trade-off between mAP and GFLOPs, outperforming state-of-the-art detectors such as YOLOv5, YOLOv8 and RT-DETR. It is noted that RT-DETR, as a transformer-based detector, may not achieve the desired accuracy on industrial defect datasets with limited amount of data. Therefore, we controlled the stability of training as much as possible on our dataset and tested the highest accuracy.

We believe that the accuracy gains across datasets mainly come from the designed dynamic cross-fusion of different models, which allows for the flexibility to extract the features required for the task from the DDPM to assist in the subsequent detection.

\subsection{Visualization of Detection Results}

Fig. \ref{heatmap} shows the visualization results for some images using GradCAM++. It can be seen that the model generates larger activation values for the important region of the defect in one of the activation maps. That is to say that the output of the model has a larger gradient value in this region, i.e., it is more sensitive to the features in this part.
In particular, it seems that there exists a certain degree of suppression in the activation of one class with respect to the other classes.

\section{Conclusion}

This paper proposes a new method to incorporate the high-order modeling capability of diffusion model into detection model to help recognize difficult targets. The proposed method uses a pre-trained denoising diffusion probabilistic model (DDPM) to extract features, and further compresses the features using a residual convolutional variational autoencoder (ResVAE). The input image is used for querying in the feature repository, which is constructed from compressed features. To fully utilize the detection model to refine the contextual features of the DDPM, we propose a dynamic cross-fusion method, where cross-attention and multi-branch self-attention are leveraged to assign weights on multiple dimensions. Finally, knowledge distillation is employed to transfer high-order modeling ability to a lightweight baseline model without additional efficiency cost. Experimental results on multiple industrial datasets demonstrate the effectiveness and efficiency of our method in improving the detection efficiency and accuracy of difficult targets. Future work will continue to explore the application of diffusion models in defect detection.

\bibliographystyle{ieeetr}
\bibliography{ref} 

\end{document}